# Diffusion Fuzzy System: Fuzzy Rule Guided Latent Multi-Path Diffusion Modeling

Hailong Yang, Te Zhang, Kup-sze Choi, *Senior Member, IEEE*, Zhaohong Deng, *Senior Member, IEEE*

*Abstract*—**Diffusion models have emerged as a leading technique for generating images due to their ability to create high-resolution and realistic images. Despite their strong performance, diffusion models still struggle in managing image collections with significant feature differences. They often fail to capture complex features and produce conflicting results. Research has attempted to address this issue by learning different regions of an image through multiple diffusion paths and then combining them. However, this approach leads to inefficient coordination among multiple paths and high computational costs. To tackle these issues, this paper presents a Diffusion Fuzzy System (DFS), a latent-space multi-path diffusion model guided by fuzzy rules. DFS offers several advantages. First, unlike traditional multi-path diffusion methods, DFS uses multiple diffusion paths, each dedicated to learning a specific class of image features. By assigning each path to a different feature type, DFS overcomes the limitations of multi-path models in capturing heterogeneous image features. Second, DFS employs rule-chain-based reasoning to dynamically steer the diffusion process and enable efficient coordination among multiple paths. Finally, DFS introduces a fuzzy membership-based latent-space compression mechanism to reduce the computational costs of multi-path diffusion effectively. We tested our method on three public datasets: LSUN Bedroom, LSUN Church, and MS COCO. The results show that DFS achieves more stable training and faster convergence than existing single-path and multi-path diffusion models. Additionally, DFS surpasses baseline models in both image quality and alignment between text and images, and also shows improved accuracy when comparing generated images to target references.**

*Index Terms*—**Diffusion Fuzzy System, Fuzzy Rule, Latent Space, Multi-Path**

## I. INTRODUCTION

In recent years, the rapid development of artificial intelligence has advanced image generation technology significantly. Emerging generative models, particularly diffusion models, have not only consistently enhanced the quality of generated images but also broadened the application domains to such areas as artistic creation [1], virtual reality [2], and medical imaging [3].

Diffusion models are currently the leading approach in image generation. Traditional single-path models, like Denoising Diffusion Probabilistic Model (DDPM) [4], generate images by adding noise in a forward process and remove it step by step in a reverse process. To reduce the high sampling cost of DDPM, Denoising Diffusion Implicit Models (DDIM) [5] introduces a non-Markovian forward process for faster sampling, but the training and inference remain expensive. Latent Diffusion Models (LDM) [6] further reduce computation by performing diffusion in a compressed latent space. Classical diffusion models often produce lower-quality images due to limited guidance during diffusion. To address this limitation, Dhariwal et al. proposed Ablated Diffusion Model with Classifier Guidance (ADM-G) [7] to integrate a classifier in the reverse process. At each step, sampling was guided by gradients from the classifier to enable class-conditional generation without retraining the model. To realize conditional generation from images, text, or multimodal inputs, Liu et al. proposed Semantic Diffusion Guidance (SDG) [8] using content and style guidance via gradients. Classifier-guided diffusion, however, is limited by the need to train the classifier and diffusion model separately. This limitation is avoided by Classifier-Free Diffusion Guidance (CFDG) [9] which uses implicit guidance, and by Guided Language to Image Diffusion for Generation and Editing (GLIDE) [10] which combines classifier-free guidance with Contrastive Language–Image Pretraining (CLIP) [11] for large-scale text-to-image generation. Unconditional Contrastive Language–Image Pretraining model (unCLIP) [12] further maps text to images for salient content generation. Despite the success, these single-path models still struggle with images having highly heterogeneous features.

Bar-Tal et al. proposed Multi-path Diffusion (MD) [13] to decompose images into subregions for parallel multi-path diffusion and fuse the outputs based on attention mechanism. Xue et al. proposed Regions Align with different text Phases in Attention Learning (RAPHAEL) [14], which further improved local details by dynamically routing heterogeneous diffusion paths based on text complexity. Residual Denoising Diffusion Model (RDDM) [15] enhances denoising robustness through residual learning. Despite these advances, multi-path models still have difficulty with globally inconsistent outputs when features vary greatly across images and focus mainly on local details. They also face challenges in inter-path coordination and higher computational cost.

his work was supported in part by the National Key R&D Program of China under Grant 2022YFE0112400, and in part by the National Natural Science Foundation of China under Grant 62176105. (Corresponding author: Zhaohong Deng).

H. Yang and Z. Deng are with the School of Artificial Intelligence and Computer Science, Jiangnan University, Wuxi 214122, China and Engineering Research Center of Intelligent Technology for Healthcare, Ministry of Education, Wuxi 214122, China. (e-mail: yanghailong@stu.jiangnan.edu.cn; dengzhaohong@jiangnan.edu.cn;).

T. Zhang is with the School of Computer Science, University of Nottingham, Nottingham, NG7 2RD, United Kingdom. (e-mail: te.zhang@nottingham.ac.uk).

K. S. Choi is with the TechCosmos Ltd., Hong Kong. (e-mail: kschoi@ieee.org).

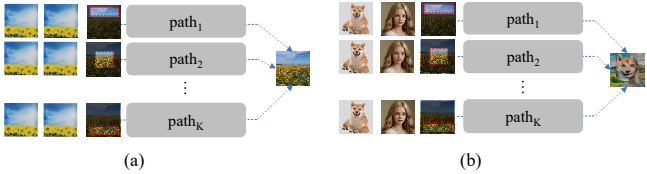

Fig. 1. Multi-path diffusion: (a) For similar images (e.g., landscapes), multi-path diffusion generates coherent images by diffusing different regions. (b) For diverse images (e.g., animals, humans and landscapes), it processes regions separately and produces outputs that are globally inconsistent.

Although multi-path diffusion models can better capture different regions than single-path models, they are not effective for image collections with large feature differences. As shown in Fig. 1(a), multi-path diffusion models can generate coherent images for similar categories (e.g., landscapes), but when dealing with images of diverse categories (e.g., animals, humans, landscapes), as shown in Fig. 1(b), they only focus on local regions and fail to model global features and produce unrealistic results. Multi-path modeling also brings challenges in inter-path coordination and higher computational cost.

Current multi-path diffusion models face three main challenges: they often capture local features only and miss global differences, struggle to coordinate multiple paths without causing mode collapse, and incur high computational costs that increase with the number of paths.

To address the limitations of existing multi-path diffusion models, this paper proposes a novel approach called the Diffusion Fuzzy System (DFS), which integrates the architectures of traditional fuzzy systems and diffusion models. Essentially, DFS can be regarded as a fuzzy-rule-guided, latent-space multi-path diffusion modeling framework. In DFS, multiple diffusion paths in latent space are built using fuzzy rules, capturing uncertainty in image diffusion.

The main contributions of this work are summarized as follows:

i) We are the first to propose the integration of fuzzy systems with diffusion models to extend traditional fuzzy system framework to diffusion learning scenarios. Diffusion Fuzzy System is proposed to implement a fuzzy-rule-guided latent-space multi-path diffusion model. It captures the uncertainty and ambiguity inherent in generative tasks explicitly, which offers a novel perspective and methodology for uncertainty modeling in generative tasks.

ii) A fuzzy rule-chain mechanism for diffusion learning is proposed to integrate cascaded fuzzy rules at each step to provide interpretable guidance. This extends the use of fuzzy rules in generative tasks and offers a new approach for guiding diffusion models.

iii) A fuzzy-membership-based latent-space compression mechanism is introduced using adaptive encoder-decoder selection to map images into low-dimensional features. This reduces computational and memory costs while ensuring accurate reconstruction. This improves the practicality of multi-path diffusion models in resource-limited or real-time scenarios.

The remainder of this paper is organized as follows. Section II introduces the fundamental concepts and principles of fuzzy systems and diffusion models. Section III presents the proposed Diffusion Fuzzy System (DFS). Section IV evaluates DFS with comprehensive experiments. Finally, Section V concludes the study and discusses future research directions.

## II. BACKGROUND

### A. Fuzzy Set and Fuzzy System

L. A. Zadeh first introduced the concept of fuzzy sets in 1965, laying the foundation for rapid development of fuzzy theory. Fuzzy set extends the classical set theory by allowing elements to belong to a set to varying degrees. Instead of an element either belonging to a set or not, the concept of fuzzy sets introduce membership function [16], [17] to quantify the degree to which an element belongs to a set. Its generalized form is expressed as follows:

$$\mu_A(x): X \to [0,1] \quad (1)$$

where $X$ denotes the universe of discourse (the set of all possible elements), $x$ is an element in $X$, and $\mu_A(x)$ represents the membership degree of $x$ in the fuzzy set $A$, with a value ranging from $[0,1]$.

For typical uncertainty inference, fuzzy system defines its input, output, and state variables using fuzzy sets. Due to their strong ability to handle uncertainty, fuzzy systems have wide application, such as automatic control, pattern recognition, and decision analysis. Currently, fuzzy systems are mainly categorized into two branches: Mamdani fuzzy systems [18] and TSK fuzzy systems [19]. Notably, TSK fuzzy systems not only offer good interpretability but also possess powerful data-driven learning capabilities, which have garnered extensive attention in recent years [20], [21].

The fuzzy rule base is the core component of a fuzzy system. Fuzzy rules are commonly used to describe the relationships between conditions and conclusions in fuzzy terms, providing the foundation for logical reasoning within fuzzy systems [22], [23]. By integrating the fuzzy rule base with components such as fuzzification, defuzzification, and the fuzzy inference engine, fuzzy system transforms fuzzy logic theory into an efficient and practical model. Fuzzy rules are typically constructed based on expert knowledge or data-driven approaches [24], [25], and are generally expressed using the "IF-THEN" format as shown below:

$$\textbf{IF } x_1 \text{ is } A_1 \text{ AND } \dots \text{ AND } x_n \text{ is } A_n, \textbf{THEN } y \text{ is } B; \quad (2)$$

where $x_1, x_2, \dots x_n$ are the input variables, $A_1, A_2, \dots, A_n$ are the fuzzy sets corresponding to the input variables (described by membership functions), $y$ is the output variable, and $B$ is the fuzzy set or precise information-processing function corresponding to the output variable.

With rapid advancement of artificial intelligence, fuzzy systems have been increasingly applied across various machine learning domains to harness their unique advantages—the interpretability of rule-based fuzzy reasoning and the powerful data-driven learning capabilities. In particular, fuzzy systems have demonstrated significant potential in areas such as transfer learning [26], [27], multi-view learning [28], [29], and clustering [30], [31].

### B. Markov Chain

Markov chain is a fundamental stochastic process widely applied in statistics, physics, computer science, and other fields [38]. A key property of Markov Chain is memorylessness, meaning that the future state depends solely on the current state and is independent of all previous states. In a Markov chain, state space refers to the set of all possible states, which can be either finite or infinite.

Hidden Markov Chain is an extension of Markov chain. It introduces the concept of hidden states [39], where true states are unobservable and hidden but can be inferred indirectly through observable data (output sequences). Based on Hidden Markov Chain, Hidden Markov Models (HMMs) are developed with two assumptions:

i) Homogeneous Markov Assumption [40]: The state transition probabilities remain constant over time, meaning that the rules do not change over time, as expressed below:

$$P(s_{t+1} = j | s_t = i) = a_{ij}, \forall t \qquad (3)$$

where the probability of transition from state $i$ to state $j$, denoted as $a_{ij}$, remains the same at any time $t$, and $s_t$ is the state variable at time $t$.

ii) Observation Independence Assumption: The observation at any given time is assumed to depend solely on the state of the Markov chain at that time and is independent of observations and states at all other times, as expressed below:

$$P(o_t | s_1, s_2, \dots, s_T, o_{t-1}, o_{t-1}, \dots, o_T) = P(o_t | s_t) \qquad (4)$$

where $o_t$ is the observation variable at time $t$, $s_t$ is the state variable at time $t$, and $t = [1, \dots, T]$.

### C. Diffusion Models

In recent years, diffusion models have achieved remarkable gradually adds random noise to the data. The model then learns the reverse diffusion process to reconstruct the desired data samples. Diffusion models simulate diffusion phenomena observed in non-equilibrium thermodynamics. The core concept is to progressively transform data from a complex distribution into a simpler one (e.g., Gaussian) by adding noise and then recovering the original data from the noise through the reverse process.

The forward diffusion process satisfies the Markov property. The initial data $x_0 \sim q(x_0)$ (from the true data distribution) is gradually corrupted by noise to obtain $\{x_1, x_2, \dots, x_T\}$, as expressed in (5). The reverse process, which denoises the data, is also modeled as a Markov chain [33]. In practice, intermediate states are predicted using AI models such as neural networks (e.g., U-Net) [34], as shown in (6), to predict the original $x_0$ from $x_t$. During training, rather than directly predicting $x_{t-1}$, the model $\epsilon_\theta(x_t, t)$ predicts the added noise $\epsilon$, as shown in (7). The training objective function for commonly used noise prediction model is given in (8).

$$q(x_t | x_{t-1}, x_{t-2}, \dots, x_0) = q(x_t | x_{t-1})$$
$$= \mathcal{N}(x_t; \sqrt{1 - \beta_t} x_{t-1}, \beta_t I) \qquad (5)$$

$$p_\theta(x_{t-1} | x_t) = \mathcal{N}(x_{t-1}; \mu_\theta(x_t, t), \Sigma_\theta(x_t, t)) \qquad (6)$$

$$\epsilon_\theta(x_t, t) \approx \epsilon \qquad (7)$$

$$\mathcal{L}_{noise} = \mathbb{E}_{x_0, \epsilon, t}[\|\epsilon - \epsilon_\theta(x_t, t)\|^2] \qquad (8)$$

where in the forward process, the data $x_0$ is gradually corrupted by adding noise, resulting in the sequence $\{x_1, x_2, \dots x_T\}$, where $x_T$ is pure noise. Here, $t = [1 \dots T]$, and $\beta_t$ is the noise variance parameter at step $t$.

## III. Diffusion Fuzzy System

### A. The Overview of The DFS Framework

As illustrated in Fig. 2, the DFS framework comprises four core modules: the Diffusion Fuzzification (DF) module, the

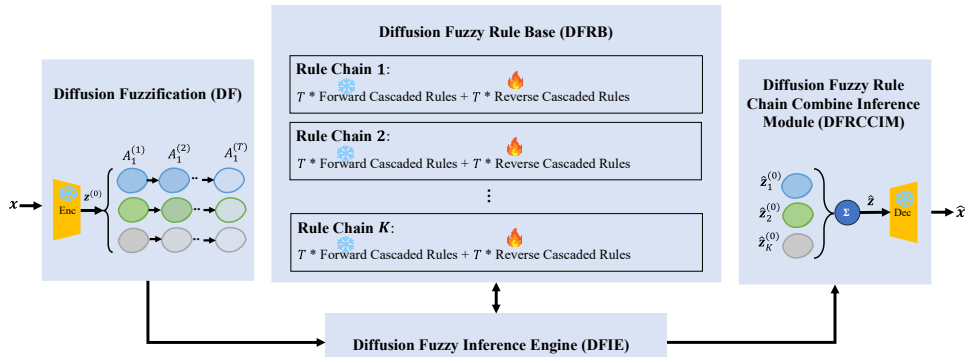

Fig. 2. DFS framework overview. It has four main modules: Diffusion Fuzzification (DF), Diffusion Fuzzy Rule Base (DFRB), Diffusion Fuzzy Inference Engine (DFIE), and Diffusion Fuzzy Rule Chain Combination Inference (DFRCCIM). DF reduces data dimensionality and computes membership degrees via the encoder Enc (orange trapezoid on the left). DFRB contains multiple rule chains with forward and backward fuzzy rules. DFIE performs rule-based inference to guide and align diffusion paths. DFRCCIM combines all rule-chain results and reconstructs high-dimensional data using the decoder Dec (orange trapezoid on the right). Blue snowflakes indicate frozen parameters, and red flames indicate learnable parameters.

success in various fields, e.g., image generation and speech synthesis [32]. Like HMMs, diffusion models are sequential models based on Markov chains. Diffusion model defines a sequence of diffusion steps within a Markov chain and

Diffusion Fuzzy Inference Engine (DFIE), the Diffusion Fuzzy Rule Base (DFRB), and the Diffusion Fuzzy Rule Chain Combination Inference Module (DFRCCIM). DFS is an innovative extension of classical fuzzy systems into the domain of generative modeling. Its primary objective is to seamlessly integrate the interpretability of fuzzy logic with the powerful generative capabilities of diffusion models. DFS extends beyond static tasks of traditional fuzzy systems by applying fuzzy inference at each diffusion step, offering interpretable guidance for generative modeling.

In the following sections, we provide a detailed description of the four key modules of DFS in the context of the text-to-image generation task.

### B. Diffusion Fuzzification

DF compresses high-dimensional data into low-dimensional features within the latent space using an encoder. It then computes the membership degrees of its intermediate states during the diffusion process—such as latent image representations, noise predictions, and feature maps—relative to different fuzzy sets, as illustrated in Fig. 3.

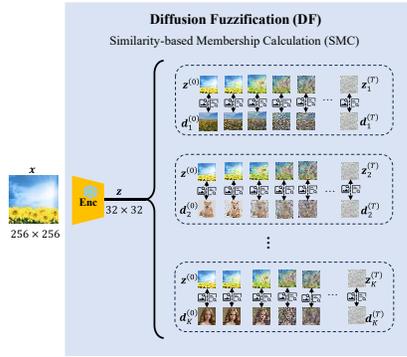

Fig. 3. Diffusion Fuzzification Module. The orange trapezoid encodes a 256×256 image into a 32×32 feature matrix. K diffusion paths (blocks in dotted-line borders) are shown on the right, with bidirectional dashed arrows indicating similarity-based membership computation (feature and semantic similarity).

In text-to-image generation tasks, when the image features within a dataset exhibit significant variation, single diffusion path model cannot achieve fitting effectively. Therefore, it is necessary to partition the dataset into multiple subsets with internally similar features and fit each subset using distinct diffusion paths.

We model each image subset as a fuzzy set, where member images are strongly associated with a particular category based on both features and semantics. The degree of membership is defined by a membership function, which reflects the fuzziness and uncertainty of the category within the image space. A representative image is selected from each subset to serve as a sample that comprehensively captures its feature-level and semantic-level characteristics. These representatives preserve the common traits of their respective subsets and can be regarded as typical instances that not only describe their corresponding subsets but also help distinguish them from others.

The fuzzification process of DF consists of the following steps: (i) representing each image subset by its representative image; (ii) performing fuzzification computations on the images.

Representative samples of an image subset can be identified in two common ways: (i) manually, based on expert knowledge, or (ii) automatically, through clustering methods. Clustering enables automatic discovery of image groups images with similar features and the selection of the most representative images from each group. Common clustering techniques include K-Means, K-Medoids, Mean-Shift, and Fuzzy C-Means (FCM) clustering.

In text-to-image generation tasks, it is often necessary to manage large-scale and diverse image datasets. For example, the MS COCO dataset contains 592,000 images of nearly 100 objects of different background categories. To efficiently identify representative real images, this study employs the K-Medoids clustering algorithm. This method selects actual data points as representatives of image subsets to achieve strong robustness against noise and outliers. Its generalized form is illustrated in the equation below:

$$\{r_1 \ldots, r_k, \ldots, r_K\} = \mathcal{C}_{medoids}(K|X) \tag{9}$$

where $X$ denotes the set of original images $x$, $K$ is the number of image subsets to be obtained, corresponding to the number of diffusion paths, $\mathcal{C}_{medoids}(.)$ is K-Medoids clustering function and $r_k$ is the representative of the $k$-th image subset.

As the diffusion process progresses, the representations of different image subsets continuously evolve, primarily reflected by varying noise levels. Since the noise injection process is predefined, representative samples from each image subset can be selected and subjected to the prescribed noise addition. This approach enables derivation of the representations of these samples throughout the entire diffusion process. Consequently, during diffusion, the representative samples of an image subset form a temporarily dynamic and sequentially correlated image sequence.

Along the same diffusion path, the subset representatives used in the membership degree computation of different rules are not the original images but the low-dimensional feature data obtained by compressing these representatives, as shown in (10). Subsequently, linear noise is added to generate the data sequence, as expressed in (11). It is important to note that the forward and reverse processes share the same representative $d_k^{(t)}$ at each corresponding step for membership computation.

$$d_k = Enc(r_k) \tag{10}$$

$$\{d_k^{(1)} \ldots, d_k^{(t)}, \ldots d_k^{(T)}\} = Noise(d_k) \tag{11}$$

where $d_k^{(t)}$ denotes the low-dimensional feature of the representative $d_k$ on the k-th diffusion path after noise has been added at step t, where $t = [1, \ldots T]$ and $T$ is the total number of diffusion steps. All diffusion paths in DFS share the same number of diffusion steps $T$. The function $Enc(.)$ is the image encoder, while $Noise(\cdot)$ is the noise injection function.

Regarding the computation of membership degrees, since image subsets are represented by their representative elements, a Similarity-based Membership Calculation (SMC) method is employed. SMC calculates the membership degree by measuring the similarity between an input image and the representative of the corresponding subset. To incorporate both feature-level and semantic-level information, as illustrated in Fig. 3, SMC determines the membership degree of an input image to a given subset by computing the feature and semantic similarities between the input image and the subset representative. This process can be formally expressed as follows:

$$SMC(\boldsymbol{z}_k^{(t)}|\boldsymbol{d}_k^{(t)}) = \alpha * \mathcal{S}(\boldsymbol{z}_k^{(t)}|\boldsymbol{d}_k^{(t)}) + (1-\alpha) \\ * \mathcal{E}(\boldsymbol{z}_k^{(t)}|\boldsymbol{d}_k^{(t)}) \tag{12}$$

where $\boldsymbol{x}$ is the input image, and $\boldsymbol{z}$ is its low-dimensional latent feature obtained via compression, i.e., $\boldsymbol{z} = Enc(\boldsymbol{x})$. $\boldsymbol{z}_k^{(t)}$ is the noisy low-dimensional feature of $\boldsymbol{z}$ at step t along the $k$-th diffusion path. $\mathcal{S}(.)$ is the semantic similarity function, $\mathcal{E}(.)$ is the feature-level similarity function, and $\alpha$ is a balancing constant between feature and semantic similarities, with a default value of 0.5.

### C. Diffusion Fuzzy Rule Combine Inference Module

The DFRCCIM module integrates the generation results from multiple diffusion paths by merging them and ultimately restoring the low-dimensional features of the images through a decoder. Common fusion strategies include weighted sum, maximum, average, and minimum methods. Considering the characteristics of diffusion-based generation scenarios, as illustrated in Fig. 4, we adopt an importance-based weighted fusion approach to combine the results from different diffusion paths. This method enables the model to focus on the diffusion modeling of different image subsets to different extents.

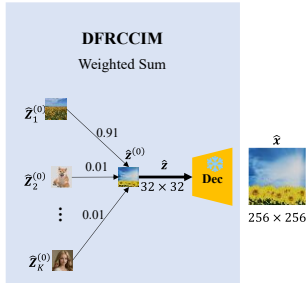

Fig. 4. Diffusion Fuzzy Rule Chain Combine Inference Module.

According to the properties of Markov chains, the fusion of diffusion paths depends solely on the outputs at the final step of each path and is independent of the intermediate-step results. Since the final step of the reverse diffusion process is $t = 1$, the multi-path fusion is performed on the outputs $\hat{\boldsymbol{z}}_k^{(0)}$ at this step. The weights are determined by the membership degrees computed from the similarity between the input image and the representatives, and the fused output is obtained via the weighted sum across all paths, as expressed in (13). Finally, DFRCCIM restores the low-dimensional fused feature back to an image using the decoder, as shown in (14).

$$\hat{\boldsymbol{z}} = \sum_{k=1}^{K} SMC(\hat{\boldsymbol{z}}_k^{(0)}|\boldsymbol{d}_k^{(0)})\,\hat{\boldsymbol{z}}_k^{(0)} \\ = \sum_{k=1}^{K} SMC(\hat{\boldsymbol{z}}_k|\boldsymbol{d}_k)\,\hat{\boldsymbol{z}}_k \tag{13}$$

$$\hat{\boldsymbol{x}} = Dec(\hat{\boldsymbol{z}}) \tag{14}$$

where $\hat{\boldsymbol{z}}_k = \hat{\boldsymbol{z}}_k^{(0)}$ is the denoised output of the $k$-th diffusion path at $t = 1$, and $\boldsymbol{d}_k = \boldsymbol{d}_k^{(0)}$ is the feature of the representative image for the image subset focused on by the $k$-th diffusion path. $Dec(.)$ is the image decoding function.

### D. Fuzzy Membership based Latent-Space Compression

To reduce the computational complexity of the multi-path diffusion process, DFS performs diffusion learning in a compressed, low-dimensional latent space. This latent-space compression has two key components: the encoder in the DF module and the decoder in the DFRCCIM module. As illustrated in Fig. 5, the original images, sized 256×256, are compressed by the encoder in the DF module into 32×32 latent features. This low-dimensional representation not only reduces data volume significantly but also mitigates the computational overhead associated with multi-path diffusion effectively.

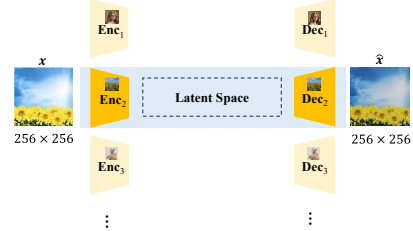

Fig. 5. Fuzzy Membership based Latent-Space Compression Mechanism: Adaptive encoder-decoder selection.

In the field of image generation, common encoder-decoder architectures include Autoencoders (AEs), Variational Autoencoders (VAEs), and Vector Quantized Variational Autoencoders (VQ-VAEs). Pretrained image encoder-decoder pairs based on these architectures are typically trained on specific datasets and image categories, which limits their generalization ability. To achieve efficient latent-space compression for datasets with highly diverse features, we integrate multiple pretrained encoder-decoder pairs. The selection of the appropriate encoder-decoder pair is determined by computing the membership degree of the input image relative to the representative features of each encoder, as expressed in (15). The encoder-decoder pair with the highest membership degree is then used to compress the input image into the latent space, as shown in (16).

$$\boldsymbol{\mu}_{AE} = SMC(\boldsymbol{R}_{AE}|\boldsymbol{x}) \tag{15}$$

$$\hat{x} = Enc_i(Dec_i(x)), i = arg\ max\ \boldsymbol{\mu}_{AE} \tag{16}$$

where $\boldsymbol{R}_{AE}$ is the array of representative encoder-decoder pairs, with $\boldsymbol{R}_{AE} \in \mathbb{R}^{M \times 3 \times 256 \times 256}$, where $M$ is the total number of encoder-decoder pairs. $\boldsymbol{\mu}_{AE}$ is the membership degree array of the input image $\boldsymbol{x}$ with respect to the encoder-decoder representatives, with $\boldsymbol{\mu}_{AE} \in \mathbb{R}^{M \times 1}$.

### E. Diffusion Fuzzy Rule Base

The DFRB module contains a generative Diffusion Fuzzy Rule Base (DFRB), as illustrated in Fig. 6. DFRB consists of multiple diffusion fuzzy rule chains, with each chain corresponding to a distinct diffusion path. Each diffusion rule chain is composed of $2 * T$ cascaded diffusion fuzzy rules. Unlike classical fuzzy rules, which typically employ linear functions in the consequent, the consequent of a single diffusion fuzzy rule is generated using a single-step diffusion or reverse-diffusion operation derived from the diffusion model.

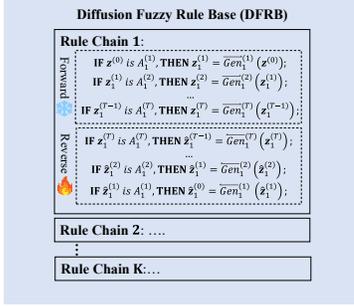

Fig. 6. Diffusion Fuzzy Rule Base. The rule base comprises multiple rule chains, each containing cascaded forward fuzzy rules and cascaded backward fuzzy rules.

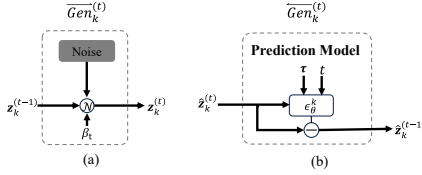

Fig. 7. Construction of the consequent of a single diffusion fuzzy rule. (a) Consequent processing operation of the forward diffusion fuzzy rule (forward generative function), (b) Consequent processing operation of the forward diffusion fuzzy rule (reverse generative function).

The temporal attributes of the rule chains in the DFRB exhibit model the dynamic evolutionary characteristics. According to the assumption of Markov chain, the system state at the current time depends solely on the state at the previous time step. In the forward diffusion process, the general form of the rule at time step t in the corresponding rule chain is given by (17), while in the reverse process, the generalized form of the rule at time step t is expressed by (18).

$$\overrightarrow{rule}_k^{(t)}: \textbf{IF } \mathbf{z}_k^{(t-1)} \textit{ is } A_k^{(t)}, \textbf{THEN } \mathbf{z}_k^{(t)} = \overrightarrow{Gen}_k^{(t)}(\mathbf{z}_k^{(t-1)}); \quad (17)$$

$$\overleftarrow{rule}_k^{(t)}: \textbf{IF } \hat{\mathbf{z}}_k^{(t)} \textit{ is } A_k^{(t)}, \textbf{THEN } \hat{\mathbf{z}}_k^{(t-1)} = \overleftarrow{Gen}_k^{(t)}(\hat{\mathbf{z}}_k^{(t)}); \quad (18)$$

where $\mathbf{z}_k^{(t)}$ is the noisy image at time step $t$ of the $k$-th diffusion path, $\hat{\mathbf{z}}_k^{(t)}$ is the denoised image at time step $t$ of the $k$-th diffusion path, $A_k^{(t)}$ is the fuzzy set at time step $t$ of the $k$-th path, $\overrightarrow{rule}_k^{(t)}$ refers to the forward fuzzy rule at time step $t$ of the $k$-th diffusion path, and $\overleftarrow{rule}_k^{(t)}$ corresponds to the reverse fuzzy rule at time step $t$ of the $k$-th diffusion path.

Fig. 7(a) illustrates the implementation of the forward rule consequent, which functions as a linear noise injection module. By modulating the temporal evolution of noise intensity, the amplitude of noise addition can be controlled, as described by the generalized formulation in (19). Fig. 7(b) depicts the data generation process in the reverse rule consequent, where the core mechanism involves predicting noise through the network $\epsilon_\theta$ and removing it from the current input, thereby recovering the lower-dimensional features from the previous time step. The generalized form of this process is presented in (20).

$$\overrightarrow{Gen}_k^{(t)}(\mathbf{z}_k^{(t-1)}) = \sqrt{1-\beta_t}\mathbf{z}_k^{(t-1)} + \sqrt{\beta_t}\boldsymbol{\epsilon} \quad (19)$$

$$\overleftarrow{Gen}_k^{(t)}(\hat{\mathbf{z}}_k^{(t)}) = \frac{1}{\sqrt{1-\beta_t}}(\hat{\mathbf{z}}_k^{(t)} - \sqrt{\beta_t}\hat{\boldsymbol{\epsilon}}) \quad (20)$$

where $\overrightarrow{Gen}_k^{(t)}$ is the consequent processing function (noise-adding function) of the rule at time step $t$ in the forward diffusion of the $k$-th path, while $\overleftarrow{Gen}_k^{(t)}$ is the consequent processing function (denoising function) of the rule at time step $t$ in the reverse diffusion of the $k$-th path. As $t$ increases, the noise intensity $\beta_t$ gradually grows, e.g., from $\beta_t = 10^{-4}$ to $\beta_t = 0.02$. $\boldsymbol{\epsilon}$ denotes Gaussian noise, and $\hat{\boldsymbol{\epsilon}}$ is the noise predicted by the network $\epsilon_\theta$.

Regarding the noise prediction network, mainstream image diffusion models (e.g., DDPM and Stable Diffusion) typically adopt U-Net-based architectures for noise prediction, which incorporates time embeddings and conditional inputs such as text or class information. To accommodate large-scale models and high-resolution image generation, some diffusion models employ Transformer-based architectures, such as the Diffusion Transformer (DiT), to enhance noise prediction capabilities. In more complex scenarios, hybrid architectures that combine U-Net and Transformer modules (e.g., Hybrid U-Net) are used to ensure semantic consistency while enabling high-resolution image generation. In the text-to-image generation task, the diffusion model in this study primarily relies on a U-Net-based design for the noise prediction network $\epsilon_\theta$. The generalized form is expressed as follows:

$$\hat{\boldsymbol{\epsilon}} = \epsilon_\theta(\hat{\mathbf{z}}_k^{(t)}, t, \boldsymbol{\tau}) \quad (21)$$

where $\epsilon_\theta$ predicts the noise $\hat{\boldsymbol{\epsilon}}$ based on the input low-dimensional feature $\mathbf{z}_k^{(t)}$, the time step t, and the condition $\boldsymbol{\tau}$.

### F. Diffusion Fuzzy Inference Engine

DFIE is the inference core of DFS. Its primary function is to perform fuzzy matching, activation, and inference over the rule chains within DFRB, based on the current fuzzified state. DFRB consists of multiple rule chains, each defining a set of diffusion fuzzy inference rules that guide an individual diffusion path. Fig. 8 presents the inference process driven by rule chains. The procedure is as follows:

**Step 1**: Cross-path alignment. Normalize the membership degrees of antecedents across the rule chains.

**Step 2**: Diffusion guidance. Apply weighting to the consequents.

**Step 3**: Pass the weighted consequents to the next cascaded rule and repeat Steps 1 and 2.

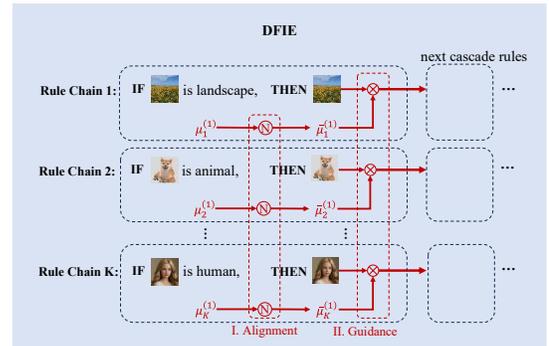

Fig. 8. Diffusion Fuzzy Inference Machine (DFIE).

Antecedent membership normalization is applied to balance the influence of different paths during rule evaluation and to prevent incoherent interactions between paths that could cause pattern collapse. Consequent weighting guides the diffusion process to focus on distinct image subsets, enabling the model to converge quickly and stably. The following sections provide detailed descriptions of Steps 1 and 2.

### 1) Antecedent Membership Normalization

To coordinate multiple diffusion paths, local cross-path alignment is applied at each step (red box in the middle of in Fig. 8, labeled "I") to stabilize rule strengths, avoid pattern collapse, and speed up convergence. During diffusion, fuzzy rules guide data generation by adjusting their weights according to activation strength to enhance or reduce their influence on image features.

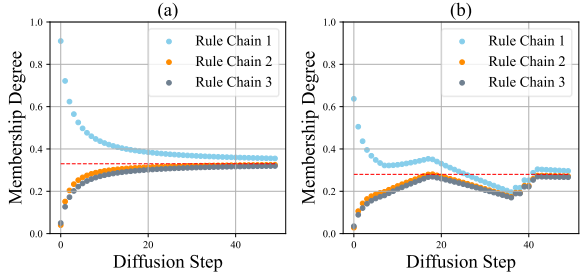

Fig. 9. Convergence of membership degrees during diffusion. (a) With alignment, membership degrees converge stably and consistently. (b) Without alignment, they fluctuate greatly and follow poor path coordination.

Multi-path alignment is achieved by normalizing rule strengths across chains at each time step, based on membership degrees from image similarity. As shown in Fig. 9(a), normalization yields smooth convergence, while Fig. 9(b) shows instability without it. The normalization forms for forward and reverse diffusion rules are given as:

$$\bar{\mu}_k^{(t)} = \overline{SMC}\left(z_k^{(t)}\big|d_k^{(t)}\right)$$
$$= \frac{SMC\left(z_k^{(t)}\big|d_k^{(t)}\right)}{\sum_{k=1}^{K} SMC\left(z_k^{(t)}\big|d_k\right)} \quad (22)$$

$$\bar{\mu}_k^{(t)'} = \overline{SMC}\left(\hat{z}_k^{(t)}\big|d_k^{(t)}\right)$$
$$= \frac{SMC\left(z_k^{(t)}\big|d_k^{(t)}\right)}{\sum_{k=1}^{K} SMC\left(z_k^{(t)}\big|d_k^{(t)}\right)} \quad (23)$$

where the normalization satisfies $\sum_{k=1}^{K} \bar{\mu}_k^{(t)} = 1$ and $\sum_{k=1}^{K} \bar{\mu}_k^{(t)'} = 1$.

### 2) Weighting of Consequent Results

DFIE guides diffusion paths by adaptively adjusting consequent weights. It reduces the weights of low-activation rules to lessen influence, and increases those of high-activation rules to strengthen feature learning.

In the forward process, noise is gradually added through rule-guided diffusion steps, each depending only on the previous state (Markov property). Outputs are weighted by rule strengths and passed to the next step as (24). The reverse process denoises the image step by step, with each result weighted by rule strengths before being fed to the preceding step as (25).

$$z_k^{(t)} = \bar{\mu}_k^{(t)} \overline{Gen}_k^{(t)}\left(z_k^{(t-1)}\right) \quad (24)$$

$$\hat{z}_k^{(t-1)} = \bar{\mu}_k^{(t)'} \overline{Gen}_k^{(t)}\left(\hat{z}_k^{(t)}\right) \quad (25)$$

Through the aforementioned mechanism, DFIE enables precise control over diffusion paths, thereby enhancing stability and robustness, and effectively addressing coordination issues among multiple paths.

### G. Training and Generation

The DFS model is trained with multi-path perceptual loss to capture diverse feature distributions. Its reverse diffusion network, guided by noise prediction, is central to both training and generation, progressively reconstructing high-quality images. This section details the training and generation strategy.

### 1) Model Training

During the training phase, a multi-path parallel training strategy is employed for DFS to capture the learning status of multi-path features and dynamically adjust learning weights. By optimizing the diffusion process through the multi-path rule mechanism, DFS progressively approaches optimal image generation performance, significantly accelerating the convergence of the diffusion model. At time step $t$ in the forward process, Gaussian noise $\epsilon^{(t)}$ is added to the image, while in the reverse process, the predicted noise $\hat{\epsilon}^{(t)}$ is obtained via the network $\epsilon_\theta$. Effective fitting of image feature data by the DFS model is achieved by minimizing the loss between the true noise $\epsilon^{(t)}$ and the predicted noise $\hat{\epsilon}^{(t)}$. Based on the multi-path parallel learning strategy, the final loss function of DFS is given in (26), and the training algorithm is detailed in Algorithm 1.

$$\mathcal{L}(\theta) = \sum_{k=1}^{K} \sum_{t=1}^{T} \bar{\mu}_k^{(t)} \mathbb{E}_{z^{(0)},\epsilon,t}\left[\left\|\epsilon^{(t)} - \epsilon_\theta\left(\hat{z}_k^{(t)}, t, \tau\right)\right\|^2\right] \quad (26)$$

where $\bar{\mu}_k^{(t)}$ is the membership degree of the input at time step $t$ along the $k$-th path, which is specifically computed using the normalized $\overline{SMC}(.)$.

---

**Algorithm 1** Training

**Inputs**: sample $x$
**Output**: the model parameters $\theta$ of $\epsilon_\theta$
1: **repeat**
2:     $z^{(0)} = Enc(x)$ // image encoding
3:     $\tau = Cond(c)$ // conditional text encoding
4:     $k = [1 \ldots K]$ // number of diffusion paths
5:     $t = [1 \ldots T]$ // diffusion path steps
6:     $\epsilon^{(t)} \sim \mathcal{N}(0, I)$ // noise added at forward step $t$
7:     take gradient descent step and guided by diffusion fuzzy rules on

$$\nabla_\theta \sum_{k=1}^{K} \sum_{t=1}^{T} \bar{\mu}_k^{(t)} \mathbb{E}_{z,\epsilon,t}\left[\left\|\epsilon^{(t)} - \epsilon_\theta\left(z_k^{(t)}, t, \tau\right)\right\|^2\right]$$

8: **until** converged
9: **return** $\theta$

---

### 2) Generating Process

The generation phase of the diffusion model relies on the reverse diffusion process, whose primary objective is to progressively reconstruct high-quality images from noisy information. In the text-to-image generation task, the conditional inputs consist of noise and textual information, while the output is an image that satisfies the given conditions.

Similarly, the input to the latent space of DFS consists of noise and text embeddings, while the output is a low-dimensional feature representation of the image. Since the noise is randomly generated, the text embedding serves as a crucial input. At each step, the model predicts the noise and progressively removes it from the data, ultimately producing a low-dimensional feature representation of the image. This representation is then restored to a normal image through a decoder. The generation algorithm of DFS is presented in Algorithm 2.

---

**Algorithm 2** Generating

**Inputs**: conditional text $c$
**Output**: generated image $\hat{x}$

1: $\epsilon \sim \mathcal{N}(0, I)$ // Gaussian noise
2: $\tau = Cond(c)$ // conditional text encoding
3: $k = [1 \ldots K]$ // number of diffusion paths
4: $t = [1 \ldots T]$ // diffusion path steps
5: **for** $t = T \ldots 1$ **do**
6:     $\hat{z}_k^{(T)} = \epsilon$ // reverse process input initialization
7:     **for** $k = 1 \ldots K$ **do**
8:         $\hat{z}_k^{(t-1)} = \overline{SMC}\left(\hat{z}_k^{(t)} \middle| d_k^{(t)}\right) \frac{1}{\sqrt{1-\beta_t}}\left(\hat{z}_k^{(t)} - \sqrt{\beta_t}\epsilon_\theta\left(\hat{z}_k^{(t)}, t, \tau\right)\right)$
9:     **end for**
10: **end for**
11: $\hat{z} = \sum_{k=1}^{K} \overline{SMC}\left(\hat{z}_k^{(0)} \middle| d_k^{(0)}\right)\hat{z}_k^{(0)}$ // path-wise result fusion
12: $\hat{x} = Dec(\hat{z})$ // image recovery from low-dimensional features
13: **return** $\hat{x}$

---

## IV. Experiments

### A. Preliminaries

#### 1) Datasets

To evaluate the performance of the proposed method, experiments were conducted using three benchmark datasets: LSUN Bedroom, LSUN Church and MS COCO [35]. Sample images are shown in Fig.10. The details of the datasets and preprocessing methods are provided in Part 1 of the Supplementary Materials section.

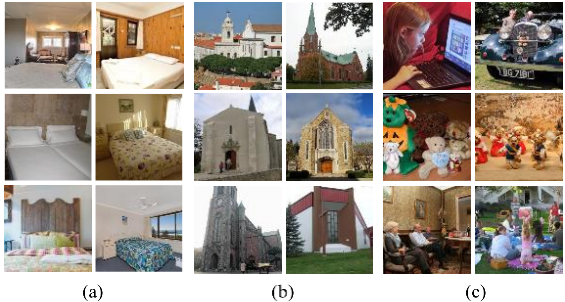

    (a)           (b)           (c)

Fig. 10. Sample images from the datasets: (a) LSUN Bedroom, (b) LSUN Church, and (c) MS COCO.

TABLE I
EXPERIMENTAL DATASETS

| Dataset | Total | Training 70% | Validation 10% | Testing 20% |
|---|---|---|---|---|
| LSUN Bedroom | 3,033,342 | 2,123,339 | 303,334 | 606,668 |
| LSUN Church | 126,527 | 88,568 | 12,652 | 25,305 |
| MS COCO | 592,000 | 414,400 | 59,200 | 118,400 |

The dataset was split into training, validation, and testing sets in a 7:1:2 ratio as shown in Table I. The training set was used for model learning, validation set for hyperparameter tuning and performance monitoring, and testing set for generalization evaluation. All experiments were repeated 10 times, and the results were reported in terms of mean and variance.

#### 2) Baselines

In the experiments, we compared the proposed DFS method with two categories of diffusion models: eight single-path diffusion models and three multi-path diffusion models. The single-path diffusion models consist of three unguided diffusion models (DDPM, DDIM, and LDM) and five guided diffusion models (ADM-G, GLIDE, CFDG, unCLIP, and SDG). The unguided models generated samples without external conditions, whereas the guided models utilized classifier guidance, text guidance, implicit classifier guidance, text-concept guidance, and semantic guidance (context and style), respectively. The three multi-path diffusion models are MD, RAPHAEL, and RDDM.

#### 3) Evaluation Metrics

In the experiments, various evaluation metrics were employed depending on the generation task. These metrics can be categorized into three types: (i) metrics used to evaluate the intrinsic quality of generated images, which include FID [36], MIFID [37], IS [38], and PSNR [39], (ii) metrics used to assess the accuracy of the generated images relative to the target images, which include SSIM [40], MS-SSIM [41], Precision [42], and Recall, and (iii) the metric used to evaluate the alignment between generated images and text captions – CLIP Score [43] (denoted as CLIP). Definitions and computational formulas for all these metrics are provided in Part 2 of the Supplementary Materials section.

#### 4) Experimental Platform and Setting

The software platform includes Python 3.8.5, Torch 2.3.0, and CUDA 12.1; the hardware platform consists of 6 NVIDIA RTX A6000 GPUs, 2 Intel 8352V CPUs, and 256 GB of memory. The settings of the DFS model are provided in Part 3 of the Supplementary Materials section.

### B. PERFORMANCE COMPARISON

This section presents the comparison experiments for image generation. Tables II–IV show the performance of the proposed DFS method compared to the baselines on the LSUN Church, LSUN Bedroom, and MS COCO datasets, respectively.

As shown in the tables, multi-path diffusion models generally outperform single-path models, with DFS achieving the best image quality across all datasets. LSUN Church and Bedroom contain relatively homogeneous scenes, while MS COCO has more diverse and complex semantics DFS also excels at image-text alignment for MS COCO. Below, we analyze DFS performance in both simple and complex semantic scenarios.

#### 1) Performance Comparison in Single-Semantic Scenarios: LSUN Bedroom and LSUN Church

The LSUN Bedroom and LSUN Church datasets are single-semantic scenarios which involve images with a single core theme, where elements like background, structure, and color are consistent. They were used to evaluate image quality and accuracy.

For LSUN Bedroom, as shown in Table II, DFS achieves the best results in six of eight metrics. For image quality, it outperforms in FID, MIFID, and IS, with PSNR 3.59 lower than MD, indicating strong overall clarity and realism. For image accuracy, DFS ranks highest in SSIM, MS-SSIM, and Precision, with Recall 0.04 lower than RDDM but still better than all single-path methods and most multi-path methods.

For LSUN Church, as shown in Table III, DFS excels at producing quality images. Particularly, it outperforms the other methods in photorealism as indicated from the FID and MIFID scores. The IS score is 1.64 lower than MD and PSNR 10.42 lower than RDDM, reflecting a focus on semantic-level over pixel-level fidelity. For structural accuracy, DFS achieves the highest SSIM and MS-SSIM. Precision and Recall are slightly lower than MD and RDDM by 0.03 and 0.05, respectively, and the impact is minimal.

Overall, DFS exhibits leading performance across multiple key metrics, effectively balancing image realism and structural accuracy.

TABLE II
PERFORMANCE COMPARISON ON THE LSUN CHURCH DATASET

| Method | FID↓ | MIFID↓ | IS↑ | PSNR↑ | SSIM↑ | MS-SSIM↑ | Precision↑ | Recall↑ |
|---|---|---|---|---|---|---|---|---|
| DDPM[4] | 7.89±0.05 | 7.19±0.08 | 13.55±0.06 | 10.24±0.03 | 0.31±0.02 | 0.17±0.02 | 0.56±0.04 | 0.51±0.07 |
| DDIM[5] | 7.76±0.06 | 7.16±0.04 | 13.65±0.04 | 11.02±0.08 | 0.34±0.06 | 0.19±0.05 | 0.59±0.04 | 0.57±0.08 |
| LDM[6] | 4.02±0.11 | 3.91±0.04 | 13.45±0.10 | 14.47±0.03 | 0.36±0.01 | 0.21±0.01 | 0.64±0.01 | 0.52±0.00 |
| ADM-G[7] | 6.74±0.03 | 6.42±0.05 | 19.23±0.05 | 19.63±0.15 | 0.41±0.03 | 0.20±0.02 | 0.61±0.04 | 0.70±0.02 |
| SDG[8] | 5.18±0.06 | 4.98±0.10 | 21.33±0.18 | 21.38±0.16 | 0.43±0.04 | 0.23±0.03 | 0.68±0.02 | 0.72±0.04 |
| CFDG[9] | 12.60±0.07 | 10.55±0.08 | 16.70±0.04 | 18.91±0.04 | 0.38±0.04 | 0.17±0.01 | 0.60±0.03 | 0.65±0.02 |
| GLIDE[10] | 13.99±0.08 | 11.25±0.12 | 15.63±0.17 | 17.56±0.14 | 0.35±0.02 | 0.18±0.03 | 0.59±0.02 | 0.61±0.04 |
| unCLIP[12] | 13.79±0.10 | 11.58±0.16 | 18.61±0.12 | 20.13±0.17 | 0.41±0.03 | 0.20±0.02 | 0.63±0.02 | 0.65±0.02 |
| MD[13] | 4.12±0.10 | 4.01±0.10 | 21.23±0.14 | **23.15±0.09** | 0.40±0.03 | 0.18±0.04 | 0.61±0.04 | 0.74±0.03 |
| RAPHAEL[14] | 4.50±0.12 | 4.32±0.13 | 22.36±0.08 | 22.68±0.19 | 0.43±0.02 | 0.21±0.03 | 0.78±0.03 | 0.78±0.04 |
| RDDM[15] | 5.98±0.06 | 5.11±0.11 | 20.51±0.11 | 20.48±0.16 | 0.39±0.02 | 0.20±0.04 | 0.67±0.02 | **0.85±0.03** |
| **DFS** | **3.81±0.03** | **3.65±0.02** | **22.80±0.09** | 19.56±0.08 | **0.45±0.02** | **0.27±0.02** | **0.79±0.06** | 0.81±0.01 |

TABLE III
PERFORMANCE COMPARISON ON THE LSUN BEDROOM DATASET

| Method | FID↓ | MIFID↓ | IS↑ | PSNR↑ | SSIM↑ | MS-SSIM↑ | Precision↑ | Recall↑ |
|---|---|---|---|---|---|---|---|---|
| DDPM[4] | 4.90±0.02 | 4.71±0.08 | 16.01±0.09 | 10.07±0.07 | 0.27±0.03 | 0.15±0.03 | 0.54±0.04 | 0.41±0.02 |
| DDIM[5] | 5.01±0.04 | 4.92±0.06 | 17.53±0.01 | 10.25±0.05 | 0.28±0.02 | 0.17±0.03 | 0.57±0.02 | 0.45±0.02 |
| LDM[6] | 2.66±0.06 | 2.61±0.02 | 17.52±0.06 | 11.55±0.02 | 0.39±0.05 | 0.20±0.04 | 0.66±0.04 | 0.48±0.03 |
| ADM-G[7] | 10.26±0.06 | 9.23±0.14 | 19.65±0.15 | 15.52±0.04 | 0.42±0.04 | 0.20±0.05 | 0.61±0.01 | 0.47±0.00 |
| SDG[8] | 5.69±0.06 | 5.02±0.12 | 20.16±0.13 | 18.69±0.15 | 0.49±0.03 | 0.22±0.03 | 0.75±0.00 | 0.69±0.02 |
| CFDG[9] | 6.23±0.07 | 5.92±0.12 | 20.36±0.17 | 17.63±0.14 | 0.45±0.00 | 0.23±0.00 | 0.70±0.06 | 0.64±0.01 |
| GLIDE[10] | 7.56±0.04 | 7.06±0.06 | 22.56±0.15 | 19.62±0.16 | 0.45±0.08 | 0.21±0.01 | 0.69±0.02 | 0.59±0.02 |
| unCLIP[12] | 6.01±0.09 | 5.50±0.18 | 21.58±0.09 | 18.36±0.06 | 0.41±0.2 | 0.20±0.04 | 0.66±0.01 | 0.58±0.01 |
| MD[13] | 2.96±0.06 | 2.86±0.14 | 23.60±0.17 | 19.65±0.15 | 0.47±0.05 | 0.19±0.02 | **0.80±0.01** | 0.72±0.02 |
| RAPHAEL[14] | 4.64±0.06 | 4.03±0.12 | 22.64±0.06 | 20.55±0.12 | 0.50±0.10 | 0.20±0.01 | 0.76±0.01 | 0.70±0.00 |
| RDDM[15] | 3.06±0.09 | 2.99±0.02 | **24.39±0.08** | **23.96±0.18** | 0.48±0.01 | 0.22±0.00 | 0.79±0.03 | **0.73±0.03** |
| **DFS** | **2.81±0.04** | **2.54±0.05** | 22.75±0.03 | 13.54±0.08 | **0.51±0.00** | **0.23±0.05** | 0.77±0.05 | 0.68±0.04 |

TABLE IV
PERFORMANCE COMPARISON ON THE MS COCO DATASET

| Method | FID↓ | MIFID↓ | IS↑ | PSNR↑ | SSIM↑ | MS-SSIM↑ | Precision↑ | Recall↑ | CLIP↑ |
|---|---|---|---|---|---|---|---|---|---|
| DDPM[4] | 32.25±0.02 | 21.03±0.9 | 8.31±0.04 | 4.24±0.13 | 0.13±0.05 | 0.09±0.02 | 0.12±0.04 | 0.12±0.03 | - |
| DDIM[5] | 34.97±0.01 | 23.01±0.02 | 4.68±0.10 | 5.17±0.02 | 0.11±0.01 | 0.07±0.01 | 0.15±0.05 | 0.08±0.01 | - |
| LDM[6] | 12.63±0.11 | 12.41±0.07 | 30.29±0.21 | 10.19±0.09 | 0.40±0.01 | 0.21±0.01 | 0.68±0.13 | 0.53±0.06 | 26.32±0.13 |
| ADM-G[7] | 10.89±0.03 | 10.23±0.06 | 15.61±0.03 | 9.78±0.02 | 0.36±0.03 | 0.17±0.01 | 0.51±0.03 | 0.61±0.05 | 20.63±0.02 |
| SDG[8] | 12.89±0.06 | 12.34±0.05 | 32.31±0.02 | 11.21±0.01 | 0.42±0.05 | 0.23±0.03 | 0.71±0.06 | 0.57±0.02 | 29.43±0.02 |
| CFDG[9] | 13.23±0.03 | 13.13±0.07 | 20.45±0.05 | 10.23±0.04 | 0.36±0.06 | 0.21±0.05 | 0.59±0.01 | 0.52±0.03 | 27.23±0.02 |
| GLIDE[10] | 7.10±0.06 | 7.08±0.07 | 16.50±0.05 | 9.91±0.04 | 0.35±0.05 | 0.21±0.04 | 0.56±0.01 | 0.63±0.03 | 28.50±0.05 |
| unCLIP[12] | 7.01±0.03 | 6.89±0.05 | 21.51±0.03 | 10.56±0.03 | 0.39±0.04 | 0.22±0.05 | 0.61±0.04 | 0.53±0.05 | 28.21±0.06 |
| MD[13] | 10.34±0.03 | 9.66±0.03 | 22.86±0.03 | 22.54±0.07 | 0.54±0.01 | 0.24±0.04 | 0.85±0.04 | 0.68±0.0.5 | 27.11±0.04 |
| RAPHAEL[14] | 6.61±0.09 | 6.5±0.01 | 30.74±0.01 | 25.60±0.07 | **0.71±0.07** | 0.30±0.09 | 0.81±0.03 | **0.72±0.02** | 29.43±0.08 |
| RDDM[15] | 13.78±0.05 | 12.52±0.07 | 28.44±0.01 | 20.72±0.09 | 0.68±0.02 | **0.32±0.02** | 0.85±0.05 | 0.51±0.02 | 29.55±0.05 |
| **DFS** | **6.29±0.07** | **6.12±0.05** | **39.15±0.01** | **26.52±0.03** | 0.48±0.03 | 0.24±0.07 | **0.89±0.01** | 0.59±0.04 | **29.61±0.06** |

\*"-" indicates no value. Since DDPM and DDIM cannot take textual conditions as input, the CLIP score for image-text alignment cannot be computed for these models.

TABLE V
ABLATION STUDY RESULTS OF DFS ON THE MS COCO DATASET

| Method | FID↓ | MIFID↓ | IS↑ | PSNR↑ | SSIM↑ | MS-SSIM↑ | Precision↑ | Recall↑ | CLIP↑ |
|---|---|---|---|---|---|---|---|---|---|
| DFS | **6.29±0.05** | **6.12±0.07** | **39.15±0.06** | **11.52±0.05** | **0.48±0.07** | **0.24±0.03** | **0.89±0.00** | **0.59±0.07** | **29.61±0.05** |
| DFS-I | 6.92±0.02 | 6.68±0.06 | 37.20±0.07 | 11.30±0.03 | 0.41±0.07 | 0.20±0.09 | 0.81±0.01 | 0.57±0.09 | 25.23±0.05 |
| DFS-IS | 7.51±0.03 | 7.23±0.02 | 32.54±0.07 | 10.95±0.06 | 0.39±0.09 | 0.16±0.03 | 0.71±0.04 | 0.54±0.07 | 21.41±0.08 |
| DFS-ISL | 7.63±0.02 | 7.31±0.05 | 32.50±0.05 | 10.80±0.01 | 0.31±0.07 | 0.15±0.04 | 0.69±0.04 | 0.54±0.01 | 21.23±0.03 |

## 2) Performance Comparison in Complex Semantic Scenarios (MS COCO)

The MS COCO dataset covers a wider range of objects and backgrounds than the LSUN Bedroom and Church datasets, which makes semantic learning more challenging. It was used to evaluate the ability of DFS in handling complex semantic scenarios.

As shown in Table IV, DFS produces outstanding image quality, with FID, MIFID, IS, and PSNR metrics outperforming all single-path and multi-path diffusion methods. Specifically, DFS achieves an FID score 0.32 lower than that of the multi-path method RAPHAEL. In terms of image accuracy, DFS attains the highest Precision, exceeding RDDM and MD by 0.04. Regarding image-to-text alignment, DFS achieves a CLIP score slightly higher than all baselines, surpassing RDDM by 0.06. This advantage arises from the multi-path diffusion approach of DFS that focuses on learning features across different image categories, where semantic constraints are embedded in the guiding rules. We designed additional prompts for the text-to-image task, and the experimental results are presented in Part 4 of the Supplementary Materials section, demonstrating the effectiveness of DFS in this task.

Overall, DFS outperforms single- and multi-path diffusion methods in image quality and accuracy for single-semantic scenarios. In complex-semantic tasks, it generates high-quality images with strong image-text alignment. Despite slight lag in some accuracy metrics, DFS demonstrates superiority in overall performance and robustness in image generation.

*C. Visualization Analysis*

The performance of four models—LDM, SDG, MD, and DFS—were visualized with random text prompts under the same conditions. Fig. 11 shows sample images from each model for an intuitive comparison of their generative performance.

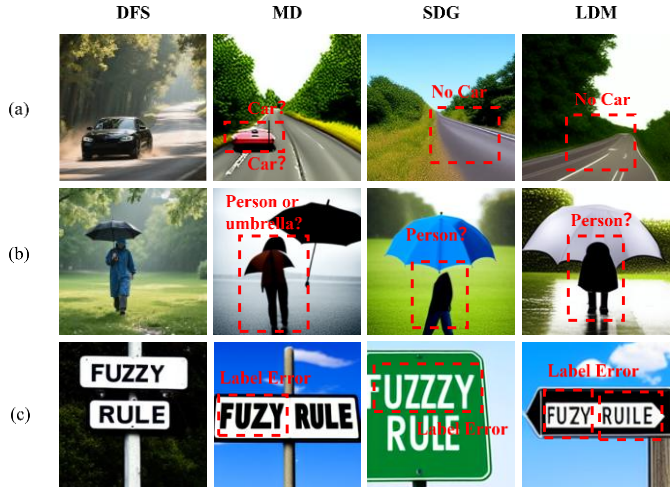

Fig. 11. Examples of images generated using the prompts: (a) *A car goes down a tree lined road*. (b) *A person holds an umbrella on a rainy day*. (c) *A street sign labeled 'Fuzzy Rule'*. The image size is 256×256 pixels.

Based on the results shown in Fig. 11, the following observations can be made:

i) For the prompt "A car goes down a tree-lined road" for Fig. 11(a), DFS and MD generate images consistent with the semantics, whereas SDG and LDM omitted elements related to the "car."

ii) For Fig. 11(b), DFS continues to generate semantically accurate images, while SDG and LDM produce elements resembling a "person" but in an incomplete form. For MD, the "person" and "umbrella" overlap, making it difficult to distinguish the "person" element.

iii) Fig. 11(c) shows that for MD and SDG, parts of the text on the "Label" are incorrect, while for LDM, all the text on the sign is wrong. Only DFS can accurately renders the content for the sign.

It is clear that MD, SDG and LDM suffer from semantic omissions and errors, whereas DFS preserves both image details and semantics, generating more coherent and commonsense-aligned images. This further demonstrates the superiority of DFS.

*D. Convergence Analysis*

To assess DFS training stability, we compared it with MD, SDG, and LDM on MS COCO. Fig. 12 shows that the convergence of model performance as training epoch increases. As shown in the figure, DFS converges fastest, reaching its inflection point at the 4th epoch and stabilizing thereafter. MD also inflects at the 4th epoch but only stabilizes after the 22nd. SDG inflects at the 5th epoch, remains stable until the 13th, then fluctuates until the 22nd. LDM inflects at the 9th epoch but continues fluctuating before the 22nd.

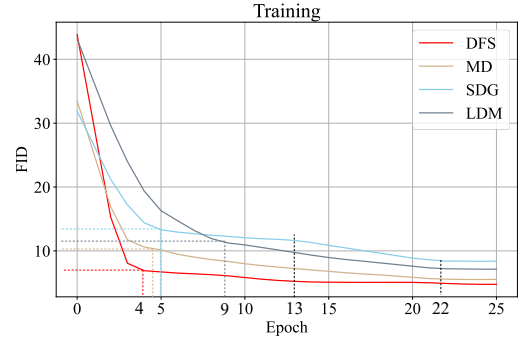

Fig. 12. Training convergence curves of DFS, MD, SDG, and LDM on the MS COCO dataset.

The MD, SDG, and LDM models only enter a stable convergence phase after more than 22 epochs, whereas DFS achieves convergence much more quickly during training.

*E. Ablation Study*

To analyze the contributions of different modules in DFS to the overall model performance, we conducted ablation experiments on the MS COCO dataset. We focused on the influence of rule chains, the DFIE engine, and latent space compression. Ablation was conducted by (i) disabling DFIE engine, denoted as DFS-I, so that rule activations were not cross-aligned and path outputs were unweighted, though the final fusion was still weighted, (ii) activating only the rule chain with the highest membership, denoted as DFS-IS, and (iii) further removing latent space compression so that diffusion was performed directly on the original images, denoted as DFS-ISL.

The ablation results in Table V show that DFS achieves the best performance, followed by DFS-I, DFS-IS, and DFS-ISL. For the pairwise comparisons DSF vs. DSF-I, DSF-I vs. DFS-IS, and DFS-IS vs. DFS-ISL, the FID score of the former is lower than those of the latter by 0.63, 0.59 and 0.12, respectively. The CLIP score of the former is higher by 4.38, 3.82 and 0.18, respectively.

From the results of the experiments, we observe:

i) Without the DFIE engine, single-rule-chain DFS-IS acts like single-path diffusion, while multi-rule-chain DFS-I acts like multi-path diffusion. The outperformance of DFS-I over DFS-IS confirms the benefit of multi-path diffusion.

ii) The DFIE engine and DFRB rule chains greatly affect DFS performance, while latent-space compression has a smaller impact, mainly reducing computational complexity and removing redundant information to facilitate model fitting.

In summary, the ablation study results demonstrate that the rule chains, DFIE engine, and latent-space compression mechanism each contribute positively to the overall performance improvement of DFS.

### F. Parameter Analysis

DFS has two key parameters: the number of steps per diffusion path and the number of diffusion paths. We analyzed their impact on MS COCO with the number of steps within 1 to 2000 and the number of paths within 1 to 10. The metrics FID, SSIM, and CLIP were used to evaluate image quality, accuracy, and text-image alignment.

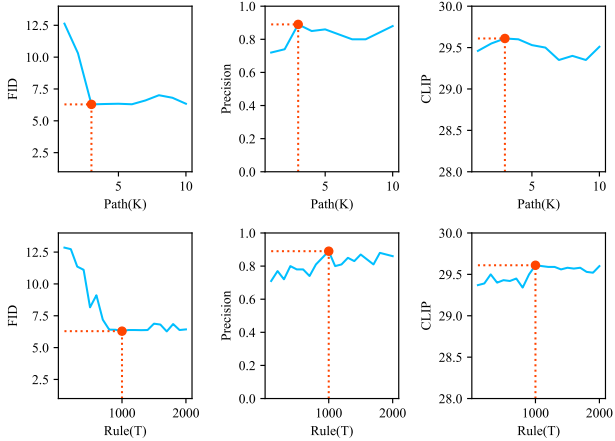

Fig. 13. Parameter analysis of DFS on the MS COCO dataset, with the number of diffusion steps ranging from 1 to 2000 and the number of diffusion paths ranging from 1 to 10.

Refer to the results in Fig. 13, with $K = 3$ diffusion paths, DFS achieves the best results: FID = 6.25, SSIM = 0.48, and CLIP = 29.01. The performance slightly declines beyond three paths but remains stable.

DFS shows multiple local optima as the number of diffusion steps increases. Overall, 1,000 steps ($T = 1000$) give the best performance with FID = 6.29, SSIM = 0.48, and CLIP = 29.33. While FID has other local optima at 1,100–1,300 steps, the other metrics peak at 1,000, making it the optimal choice.

In Part 5 of the Supplementary Materials, , we analyze DFS parameters on LSUN Bedroom and Church. For simple semantics, CLIP scores are unreliable and FID and Precision are used. DFS performs best with 3 diffusion paths and 1,000 steps per path.

Overall, considering all the three datasets, DFS achieves locally optimal performance across all metrics when the number of paths is 3 and each path contains 1,000 diffusion steps.

### G. Statistical Test

This section conducts statistical tests on the FID performance of DFS and the eight baseline methods, including the a priori Friedman test [44] and the post hoc Holm test [45].

Using the Friedman test, p-values were calculated to assess statistical differences among the methods. Table VI shows the results of DFS and the eight baselines. DFS achieves the best average rank. The test statistic is 27.25 with 8 degrees of freedom, yielding a p-value of 0.00064, which is well below 0.05, indicating that the performance of DFS is significantly different from the methods.

TABLE VI
P-VALUES AND RANKINGS OF THE 9 METHODS

| Algorithm | Ranking | statistic | p-value |
|---|---|---|---|
| DFS | 1.25 | | |
| DDPM | 8.75 | | |
| DDIM | 7.75 | | |
| ADM-G | 5.375 | | |
| GLIDE | 3.5 | 27.25 | 0.00064 |
| CFDG | 6.5 | | |
| unCLIP | 4 | | |
| LDM | 2 | | |
| SDG | 5.875 | | |

TABLE VII
HOLM POST-HOC COMPARISON ($\alpha = 0.05$)

| $i$ | vs. DFS | $z = (R_h - R_i)/SE$ | p-value | Holm | Hypothesis |
|---|---|---|---|---|---|
| 8 | DDPM | 3.872983 | 0.000012 | 0.006250 | Reject |
| 7 | DDIM | 3.356586 | 0.000079 | 0.007143 | Reject |
| 6 | CFDG | 2.711088 | 0.000171 | 0.008333 | Reject |
| 5 | SDG | 2.38834 | 0.000625 | 0.010000 | Reject |
| 4 | ADM-G | 2.130141 | 0.000131 | 0.012500 | Reject |
| 3 | unCLIP | 1.420094 | 0.000958 | 0.016667 | Reject |
| 2 | GLIDE | 1.161895 | 0.002127 | 0.025000 | Reject |
| 1 | LDM | 0.387298 | 0.009451 | 0.050000 | Reject |

Pairwise comparisons between DFS and the eight baselines were then conducted using the post-hoc Holm test. The results in Table VII show that the null hypotheses that there is no statistically significant difference in performance between the pairs are all rejected, indicating DFS significantly outperforms all the baselines.

### H. Computational Resource and Complexity

The resource consumption and complexity of DFS was evaluated in comparison with the baselines MD, SDG and LDM. The details are discussed in Part 6 of the Supplementary Materials section. Overall, DFS exhibits a clear advantage among multi-path-based approaches.

## V. CONCLUSION

In generative AI, diffusion models are widely used for image generation and have greatly improved image quality. Yet, they still face challenges due to images with diverse features, difficulty in coordinating multiple diffusion paths, and high computational cost. To address these issues, we propose the Diffusion Fuzzy System. DFS constructs multiple diffusion paths based on rule chains, enabling effective modeling of image sets with diverse features and thereby improving the quality of generated images. DFS introduces a rule inference engine to perform reasoning, rule alignment and weighted aggregation, which resolves coordination issues among multiple diffusion paths. Experiments on LSUN Bedroom, LSUN Church, and MS COCO show that DFS outperforms both single-path models and multi-path models in image quality, text-image alignment, and target accuracy. Despite its performance, DFS still has limitations. Currently, all the paths have the same number of steps, which restricts multi-scale feature learning. Moreover, although latent compression reduces overhead, the growing number of rules increases spatial complexity and limits scalability. Future work will focus on: (i) allowing variable-length diffusion paths for richer feature learning, (ii) detecting and sharing redundant steps across paths to lower complexity, and (iii) introducing scene-aware dynamic path activation to enhance adaptability and efficiency. These improvements are expected to further strengthen DFS's generative power and applicability.